\def\ProofPathsArxivVersion{1}
\documentclass[conference,letterpaper]{IEEEtran}
\IEEEoverridecommandlockouts
\usepackage[T1]{fontenc}
\usepackage{amsmath,amssymb}
\usepackage{graphicx,booktabs}
\usepackage{tikz}
\usetikzlibrary{arrows.meta}
\usepackage[nocompress]{cite}

\usepackage[hidelinks]{hyperref}
\definecolor{ppblue}{HTML}{244A73}
\definecolor{ppteal}{HTML}{167465}
\definecolor{ppgold}{HTML}{946418}
\ifdefined\ProofPathsArxivVersion
\usepackage{eso-pic}
\AddToShipoutPictureFG*{%
\AtPageLowerLeft{%
\raisebox{0.2in}[0pt][0pt]{%
\makebox[\paperwidth][c]{%
\parbox[b]{\textwidth}{\normalfont\scriptsize
\copyright\ 2026 IEEE. Permission from IEEE must be obtained for all other uses, in any current or future media, including reprinting/republishing this material for advertising or promotional purposes, creating new collective works, for resale or redistribution to servers or lists, or reuse of any copyrighted component of this work in other works.
}}}}}
\fi
\hypersetup{pdftitle={ProofPaths: An AI Agent Framework for Assessing Student Reasoning},
  pdfauthor={Mohsen Hariri and Vipin Chaudhary},
  pdfsubject={Educational assessment and software architecture},
  pdfkeywords={ProofPaths, educational assessment, student interviews, HPC inference}}

\title{ProofPaths: An AI Agent Framework for\\Assessing Student Reasoning}
\author{
\IEEEauthorblockN{Mohsen Hariri}\thanks{This work was supported in part by National Science Foundation (NSF) awards 2117439 and 2320952.}
\IEEEauthorblockA{Case Western Reserve University\\
Cleveland, Ohio, USA\\
mohsen.hariri@case.edu}
\and
\IEEEauthorblockN{Vipin Chaudhary}
\IEEEauthorblockA{Case Western Reserve University\\
Cleveland, Ohio, USA\\
vipin@case.edu}
}

\IEEEaftertitletext{%
{\centering
\begin{tabular}{@{}c@{\hspace{8pt}}l@{}}
\begin{minipage}[c]{12mm}
\centering
\includegraphics[width=12mm]{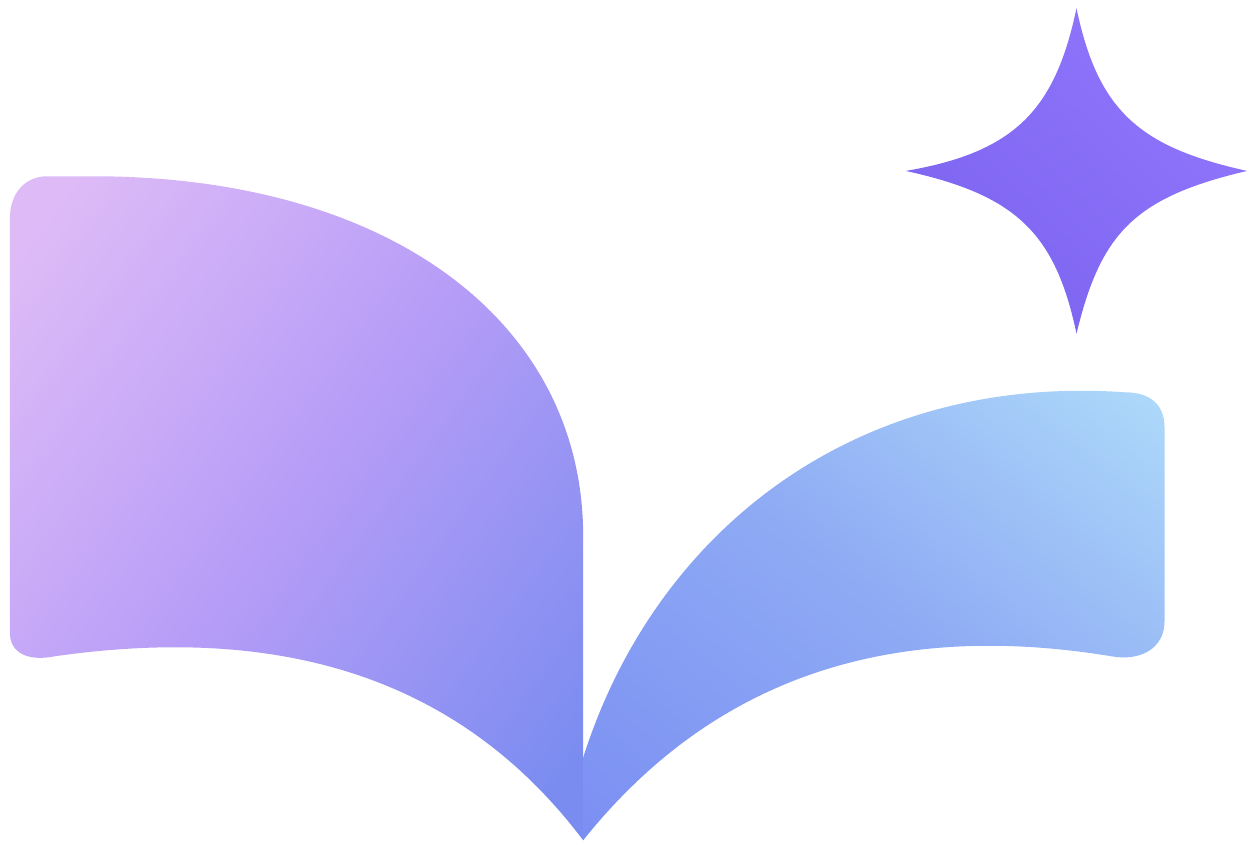}
\end{minipage}&
\begin{tabular}{@{}c@{}}
{\normalsize\bfseries\url{https://proofpaths.github.io}}\\[2pt]
{\small\href{mailto:proofpaths@case.edu}{\nolinkurl{proofpaths@case.edu}}}
\end{tabular}
\end{tabular}
\ifdefined\ProofPathsArxivVersion
\par\smallskip
{\small Accepted at the 8th Workshop on Education for High Performance Computing (EduHiPC 2026).\par}
\smallskip
{\small If you are interested in using ProofPaths, \href{mailto:proofpaths@case.edu}{contact us}.\par}
\fi
\par}
}

\begin{document}
\bstctlcite{IEEEcontrol}
\maketitle

\begin{abstract}
Students can use large language models to complete assignments, but a correct solution does not establish whether they understand the reasoning behind it. Individualized interviews can probe that understanding but require recurring staff effort. We introduce ProofPaths, an AI agentic framework that combines rubric-based grading with individualized and interactive interviews. Multiple grading agents assess a submission against instructor guidance, exchange critiques, and revise their judgments. Their scores determine which review supplies the grade and feedback. An interview agent asks students to justify steps in their submission and uses their answers to guide follow-up questions. Each question is linked to a retrieved submission excerpt, so instructors can compare the student's explanation with the cited passage. The system tracks assessment progress and stores grading reviews and interview transcripts for each submission. Several deterministic tests of the assessment controllers check source handling, score aggregation, and limits on correction attempts. Instructors can inspect all reviews and transcripts and override grades within the assignment workflow.
\end{abstract}

\begin{IEEEkeywords}
educational assessment, large language models, student interviews, software architecture, HPC inference
\end{IEEEkeywords}

\section{Introduction}
Large language models (LLMs) are used for programming assistance and tutoring~\cite{codeaid2024}, \cite{kestin2025tutoring}. Their effects on learning depend on how that assistance is provided. In an undergraduate physics trial, Kestin et al. found higher immediate post-test scores after a structured AI tutoring lesson than after an in-class active-learning lesson~\cite{kestin2025tutoring}. In high-school mathematics, Bastani et al. found that unrestricted GPT-4 assistance improved practice performance but reduced subsequent unaided exam performance, whereas teacher-designed hints mitigated the harm~\cite{bastani2025learning}. A correct AI-assisted solution alone does not establish what a student understands.

A student can, for example, submit the correct normal-equation solution for least squares yet believe that every quadratic loss has a unique minimizer. A follow-up about a duplicated feature tests the assumption behind the calculation: uniqueness requires linearly independent columns in the feature matrix, and duplicating a feature removes that independence~\cite{boyd2018applied}. For AI-assisted work, the student's explanation provides evidence of whether they understand the reasoning behind the AI's answer or guidance.

Chi et al. associated successful problem solving with students' explanations of how solution steps relate to principles and when those steps apply~\cite{chi1989self}. Kannam et al. used interviews led by teaching assistants (TAs) to ask students to explain and extend their take-home programming work. Their implementation used three TAs per thirty-student section, a staffing level they reported as unsustainable without changes to TA responsibilities~\cite{kannam2025}.

\begin{figure*}[t]
\centering
\resizebox{\textwidth}{!}{\input{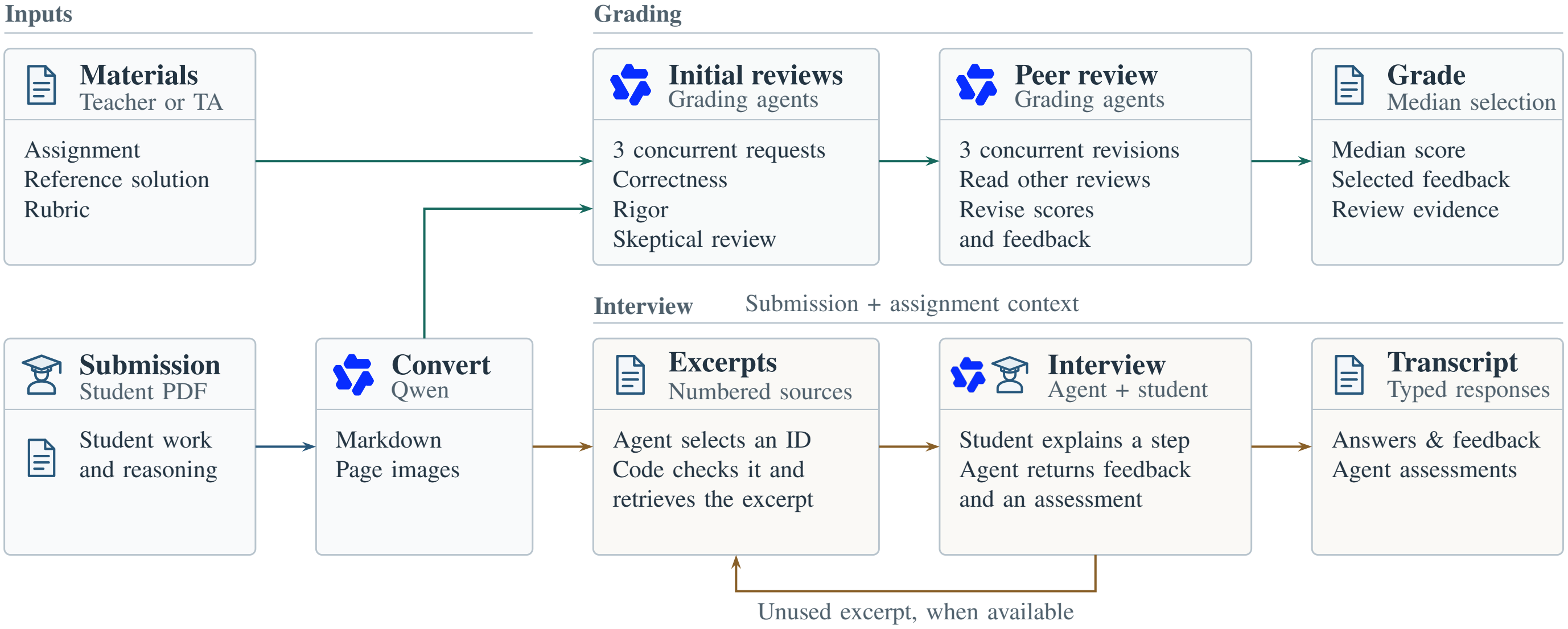}}
\caption{Grading and interviewing use the same submission. Grading agents exchange critiques and revise their assessments before one review supplies the grade and feedback. The interview agent asks about submission excerpts and follows up on the student's answers. The interview proceeds independently of grading.}
\label{fig:workflow}
\end{figure*}

We introduce ProofPaths, a framework that uses LLM agents for rubric-based grading and individualized interviews about students' solutions (Fig.~\ref{fig:workflow}). Teachers or TAs provide the assignment, reference solution, and grading criteria. Multiple grading agents assess each submission against these materials, exchange critiques, and revise their scores and feedback. An interview agent asks students to justify steps in their work and uses their answers to choose follow-up questions. Each question is linked to a retrieved passage from the submission. Teachers and TAs can inspect the grading reviews and interview transcript and override grades.

The system tracks assessment progress and records which inputs produced each result, including when requests are retried or students replace their work. Institutional high-performance computing (HPC) resources provide model inference. Deterministic tests with scripted agent responses check source retrieval, score aggregation, and limits on correction attempts.

\section{Related work}
Gradescope supports shared rubrics for grading handwritten assignments~\cite{gradescope2017}, and Policar et al. evaluated LLM-generated grades and feedback in a bioinformatics course~\cite{policar2025}. ProofPaths extends submission assessment with questions about the decisions in the work, so an instructor can inspect both the evaluation and the student's explanation.

Kannam et al.'s code interviews asked students to explain and extend completed programming assignments in exchanges with teaching assistants~\cite{kannam2025}. ProofPaths implements this relationship between assignment and interview through an LLM agent, typed responses, and explicit source excerpts. The shared assessment purpose motivates the design; the changed interviewer and response format require educational validation.

CodeAid supports students while they are programming, using conceptual explanations and pseudocode while avoiding direct code solutions~\cite{codeaid2024}. ProofPaths instead starts from completed work and asks the student to account for its reasoning. Executable assessment offers another form of evidence: CodeRunner checks programs against tests~\cite{coderunner2016}, and Gradescope's autograder runs instructor-supplied grading code~\cite{gradescopeautograder}. ProofPaths elicits explanations of student work; it has no program-execution stage. Its contribution lies in integrating rubric review, individualized questioning, and an inspectable record within one assignment workflow.

\section{Assessment workflows}
\label{sec:design}
Instructors prepare the assignment, reference solution, and grading instructions, with optional assistance from an AI agent. Students submit their solutions and answer interview questions in writing. Grading assigns credit using the instructor's criteria; interviewing elicits explanations of the assumptions and decisions in the solution.

An assessment skill combines role instructions with selected input materials and a validated response schema. Grading receives the reference solution and rubric; interviewing uses the submission and prior answers. Agents can be configured to use the same model or different models, with controller checks governing progression.

\subsection{Grading with review and revision}
Let $x$ be the parsed student solution and its page images, $a$ the assignment, $r$ the reference solution, and $b$ the grading instructions. Each grading agent receives $(x,a,r,b)$. Three agents examine correctness, rigor of reasoning, and the risk of awarding unwarranted credit, respectively. Each returns an overall score, criterion scores, feedback, and a confidence estimate. Their initial reviews run concurrently. In a second concurrent round, each agent reads the other reviews and revises its assessment.

The assessment controller validates score ranges and requires a common positive maximum $M$ for the revised evaluations. A scale mismatch triggers bounded correction. For revised scores $s_1,s_2,s_3$, the automated grade is
\begin{equation}
 s^{\star}=\operatorname{median}(s_1,s_2,s_3),\qquad 0\leq s_i\leq M.
\label{eq:median}
\end{equation}
The review with the median score supplies the summary, feedback, and criterion scores; stable ordering resolves ties. Selecting an existing review preserves the attribution of those fields to the judgment that produced them. The application retains the other reviews as well. If two scores lie in an interval $[L,U]$, the median with any third score remains in $[L,U]$, bounding the influence of one numerical outlier. ProofPaths reports the average of the grading agents' confidence estimates.

\subsection{Interviews tailored to student work}
The interview agent is instructed to ask about a claim, equation, or decision in the student's solution. Each request includes $x$, assignment context $a$, and prior exchanges, with no designated fields for $r$, $b$, or grading results. The student's answer becomes context for the next question. This makes a misunderstanding expressed during the interview available for follow-up, even when the step is correct. Assignment credit comes from grading; the interview retains explanations and feedback for separate instructor review.

Each question is associated with a numbered excerpt from the prepared set $E(x)=\{(i,e_i)\}_{i=1}^{m}$. Extraction considers sentences before their containing paragraph, followed by a pass over lines. Candidates are deduplicated after whitespace and case normalization, truncated at a word boundary where possible, and capped at 24 excerpts. For each question $q_t$, the agent returns question text and an excerpt identifier $i_t$. The controller validates the identifier and retrieves the source:
\begin{equation}
 i_t\in\{1,\ldots,m\},\qquad \text{source}(q_t):=e_{i_t}.
\label{eq:source}
\end{equation}
Retrieval by identifier preserves the prepared source text, including notation. An excerpt cannot be reused while an unused one remains, encouraging questions to cover additional passages. Before advancing, the controller checks that earlier excerpts still occur as normalized substrings of the submission. These rules keep the source association explicit throughout the exchange.

\begin{figure}[t]
\centering
\resizebox{\columnwidth}{!}{\input{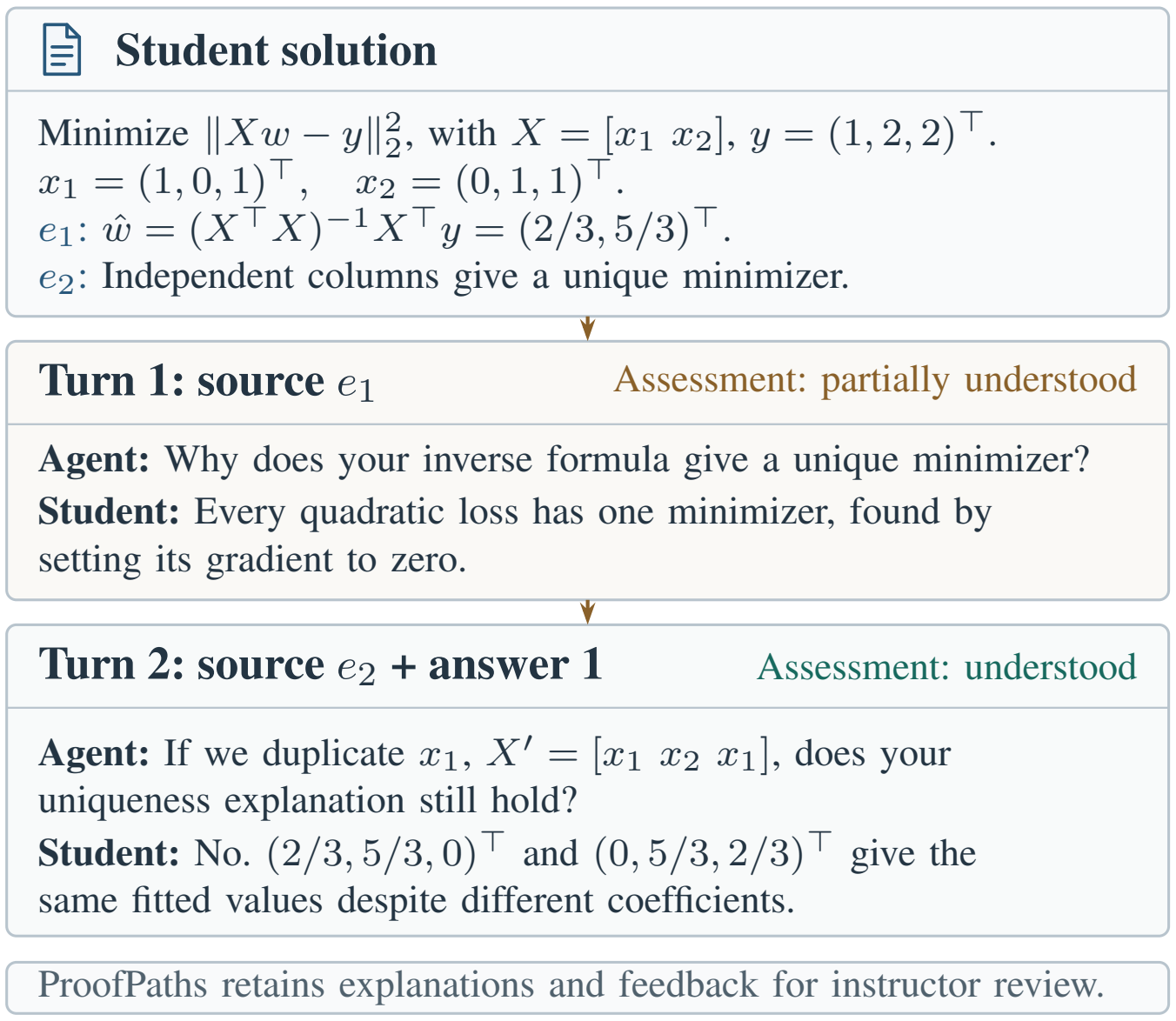}}
\caption{A least-squares interview probes the assumption behind a correct calculation. The student's first explanation overlooks the rank condition; the follow-up changes the feature matrix to test that explanation. The exchange and understanding labels are authored illustrations.}
\label{fig:example}
\end{figure}

In Fig.~\ref{fig:example}, the first explanation attributes uniqueness to the quadratic loss. Duplicating a feature produces distinct coefficient vectors with the same fitted values, testing recognition of the missing rank condition~\cite{boyd2018applied}. The second question uses a new excerpt and the earlier answer to examine the justification under a changed assumption.

The interview agent provides feedback and labels each answer as understood, partially understood, or not understood; it also provides closing feedback after the final answer. A $Q$-question interview uses one inference call for the opening question and $Q$ further calls to process answers and generate subsequent questions or closing feedback.

ProofPaths uses multidimensional item response theory (MIRT) to estimate student ability and guide follow-up selection. The model is implemented with Scorio\footnote{\url{https://github.com/mohsenhariri/scorio}}, a library developed for model evaluation~\cite{hariri2026dont}, \cite{hariri2026ranking}. Selected questions remain subject to the same excerpt checks.

\subsection{Validation and question delivery}
Separate schemas for starting, continuing, and completing an interview reject extra fields, invalid score ranges, and invalid states before the exchange advances. A schema or controller error permits two correction attempts before explicit failure. The transport client can also retry malformed JSON, so three workflow attempts may involve more than three inference requests.

Questions and excerpts support image-based and accessible presentation. The browser displays a watermark, requests fullscreen where supported, and records focus, visibility, and specified capture-related events. ProofPaths can invalidate an attempt when the browser reports a violation; students can then restart. Staff can moderate submissions and override grades through the application.

\section{Framework architecture}
\label{sec:architecture}
\begin{figure*}[t]
\centering
\resizebox{\textwidth}{!}{\input{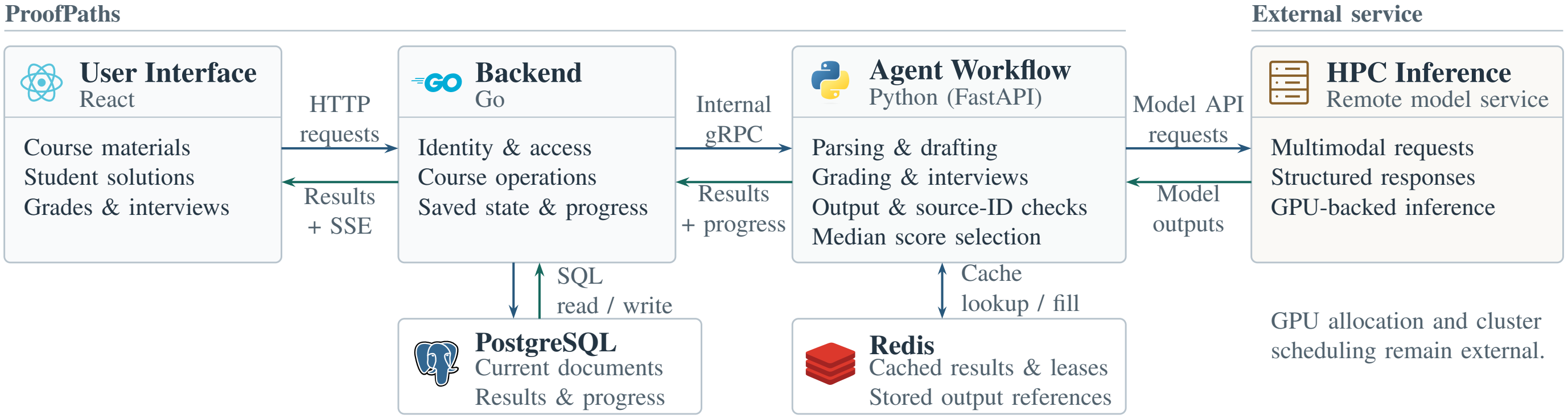}}
\caption{Course access and persistent assessment state are managed by Go and PostgreSQL. Python coordinates parsing and agent workflows, with Redis for completed outputs and work leases. Institutional HPC resources serve model inference. Results and stored progress reach the browser through Go.}
\label{fig:architecture}
\end{figure*}

Course administration and assessment use separate services; an HPC service supplies model inference (Fig.~\ref{fig:architecture}). Persistent storage retains assessment inputs and progress across document processing, model requests, and student exchanges.

\subsection{Application, workflow, and inference services}
The public Go backend authenticates users, checks course enrollment and permissions, and enforces assignment availability. PostgreSQL stores course materials, submissions, and assessment records. A Python service built with FastAPI parses student solutions and coordinates the agents, returning results to Go for storage and delivery. The Python and inference services remain behind the public application interface, with network access controlled by the deployment.

An OpenAPI specification defines the browser interface and generates TypeScript types. Protocol Buffers define internal calls between Go and Python. Runtime validation checks agent responses for structure, score ranges, source identifiers, and permitted state transitions. Parsing and inference use replaceable clients, allowing controller behavior to be tested with scripted responses.

The parsing pipeline produces Markdown from student solutions. Assessment requests include the parsed text and source images so agents can inspect notation and diagrams. The assignment and instructor materials supply the disciplinary context.

The model client uses an OpenAI-compatible interface with vLLM-compatible generation and structured-output options. Institutional NVIDIA A100 GPUs serve inference, with key--value (KV) cache quantization to reduce cache memory requirements~\cite{hariri2026quantize}. The inference service manages GPU allocation and scheduling.

\subsection{Assessment latency and inference cost}
Initial grading reviews can run concurrently, but peer revision waits for the first round to finish. Interview questions depend on preceding answers and therefore advance sequentially. For $P$ parsing calls, $G$ grading agents, and $Q$ interview questions, the total is $P+2G+Q+1$ model requests: $P$ for parsing, $2G$ for grading, and $Q+1$ for the interview. This count assumes no cache reuse and excludes correction calls, transport retries, and optional material drafting.

Let $c_p$ be the duration of parsing call $p$ and $g_j^{(r)}$ the duration of grading agent $j$'s call in round $r$. With no correction calls or retries, the uncached time from the start of parsing to a grade is
\begin{equation}
 T_{\mathrm{grade}}=\sum_{p=1}^{P}c_p+
 \max_{1\leq j\leq G}g_j^{(0)}+\max_{1\leq j\leq G}g_j^{(1)}+\delta,
\label{eq:critical-path}
\end{equation}
where $\delta$ accounts for rendering, persistence, and controller overhead outside these calls. The maxima identify the slowest review in each round, and the sum captures sequential parsing. Call durations include contention at the shared inference endpoint, so concurrency alone supplies no fixed speedup. The interview can start after parsing while grading is still running; its duration includes $Q+1$ sequential inference calls and student response time. Thus, document length determines conversion demand, while interview length adds sequential work for each student.

\subsection{Course materials and assessment records}
Staff prepare course materials with optional AI drafting assistance. ProofPaths places generated descriptions and grading instructions in editable forms and stores generated reference solutions with the assignment. Each assessment record links the student's solution and parsed content to the grading reviews and interview transcript.

An immutable manifest identifies each operation's inputs and configuration. Replacing a submission or restarting an interview changes the active attempt version. Go stores parsing, grading, and interview status separately and persists streamed grading updates. It polls for stored parsing and grading progress and sends updates to the browser through server-sent events (SSE). The manifest and attempt version determine which results belong to the active assessment.

\section{Distributed assessment execution}
\label{sec:distributed}
If a student replaces a solution during assessment, earlier results must remain linked to the inputs used to generate them. Retries can also trigger repeated inference for the same request. ProofPaths uses Redis for reusable outputs, PostgreSQL for result acceptance, and gRPC for communication between Go and Python.

\subsection{Reusing results}
ProofPaths reuses cached results when their inputs, settings, and access scope match the request. For workflow stage $z$, the cache key is
\begin{equation}
 K_z=H\bigl(\operatorname{encode}(\sigma,z,v_z,d_z,c_z)\bigr),
\label{eq:cache-key}
\end{equation}
where $H$ is a cryptographic digest and $\operatorname{encode}$ is an unambiguous canonical encoding. The tuple includes access scope $\sigma$ at the institution/course/user level, stage and output-schema version $v_z$, ordered input digests $d_z$, and stage-specific configuration $c_z$. Rendering keys include document bytes and resolution; transcription keys also include converter instructions, model revision, and generation settings. Redis stores bounded metadata and references to completed outputs, with page images in persistent storage. Authorization precedes cache lookup.

Assessment keys include the attempt version, relevant teaching-material digests, reviewer role, interview history where applicable, and run identifier. Reviewer roles and experimental replicates therefore remain distinct logical operations. Only outputs that pass schema and controller checks enter the completed cache. Changing a submission, rubric, or correction instruction selects a new key. Expiration controls retention; matching the key controls reuse. Reusing a transcription retains an earlier model output rather than assuming that a fresh call would reproduce it.

A worker takes a short lease on an entry being filled to limit duplicate computation. Redis \texttt{SET} with \texttt{NX} and \texttt{PX} supports conditional insertion and expiry~\cite{redisset}; release atomically checks the owner's token before deletion. Because lease expiry, eviction, or failover can permit duplicate execution, ProofPaths enforces result acceptance separately. Before dispatch, PostgreSQL records the operation identifier and binds it to the immutable manifest. ProofPaths rejects reuse of an operation identifier with conflicting inputs. A transaction accepts a result only if its manifest and attempt version match the active operation and no result has committed under that identifier. This permits at most one accepted result per logical operation even when inference runs more than once.

\subsection{Recovering progress and limiting work}
gRPC generates interfaces from Protocol Buffers definitions and supports streamed server responses~\cite{grpcconcepts}. Conversion, grading, and interview request types carry an attempt version, operation identifier, and input digests. The interview type excludes designated reference-solution and rubric fields. Grading progress events carry an operation identifier and monotonically increasing sequence number. Go saves each event before forwarding it to the browser, so the browser can retrieve stored progress after reconnecting.

Explicit deadlines bound each call. Python passes the remaining budget to inference and stops spawned work on cancellation, since deadline expiry alone does not clean up application work~\cite{grpcdeadlines}. Admission control limits outstanding inference calls. Transport retries preserve the operation identifier and use bounded backoff; they are separate from requests to correct an agent's output. Once response headers are received, gRPC commits the call and stops automatic retries~\cite{grpcretry}. Recovery of an interrupted progress stream therefore uses saved application state rather than replaying the grading workflow.

\section{Functional validation}
\label{sec:validation}
\subsection{Inspecting the reasoning behind a solution}
The authored exchange in Fig.~\ref{fig:example} retains a correct calculation alongside an overgeneralized explanation and a response to a changed assumption. A solution score alone would omit that disagreement. Initial and revised grading reviews separately record how peer critiques affect an agent's judgment. Median selection preserves one review's score and feedback. These records expose the basis of assessment.

\subsection{Controller behavior}
We tested the controllers with nine deterministic probes using a synthetic algebra solution, three prepared excerpts, and scripted agent responses. The controllers behaved as specified in all nine probes (Table~\ref{tab:validation}). The probes isolate the rules governing source handling, aggregation, and correction.

\begin{table}[t]
\caption{Controller behavior in nine deterministic probes with synthetic inputs and scripted agent responses. Acceptance denotes passing the implemented checks.}
\label{tab:validation}
\centering
\footnotesize
\begin{tabular}{@{}p{0.68\columnwidth}p{0.27\columnwidth}@{}}
\toprule
Input or perturbation & Observed behavior \\
\midrule
Valid excerpt identifier & Exact excerpt \\
Unknown excerpt identifier & Rejected \\
Reuse with unused excerpts & Rejected \\
Reuse after excerpt exhaustion & Accepted \\
Altered stored source text & Rejected \\
Unrelated question, valid identifier & Accepted \\
Criterion/overall inconsistency & Accepted \\
Upper score: 95 to 100 & Median stays 82 \\
Three invalid interview outputs & Explicit failure \\
\bottomrule
\end{tabular}

\end{table}

The controller returned the exact prepared excerpt for a valid identifier and rejected unknown identifiers. It rejected altered historical source text before making another inference request. It also rejected excerpt reuse while unused excerpts remained and permitted reuse after exhaustion.

For scores $(80,82,95)$, increasing the highest score to 100 retained a median of 82, confirming the aggregation rule's response to that perturbation. The controller raised an explicit error after three successive invalid interview outputs exhausted the workflow attempts. The controller accepted an unrelated question with a valid identifier and a review with inconsistent criterion and overall scores.

\section{Discussion and limitations}
ProofPaths individualizes interviews through the student's solution and earlier answers. Instructors prepare reference solutions and grading criteria at the assignment level; task-specific skills guide rubric review and follow-up questioning for each student. This division of responsibilities automates recurring assessment tasks while keeping criteria and final grades under instructor control. Retained reviews make grading judgments inspectable, and the interview transcript links each question to its source excerpt and the student's response.

Controllers enforce response structure, excerpt membership, and score bounds. The synthetic probes in Table~\ref{tab:validation} verify these rules using prepared agent responses. Question relevance and consistency between criterion scores and overall scores are outside these checks; reviewers are also not required to use identical criterion names. Instructors can compare each question with its cited passage and each overall score with the recorded criterion scores.

Peer revision exposes changes in judgment, and the median bounds the influence of one numerical outlier. Because agents exchange critiques, agreement among their revised scores is not independent corroboration, even when they use different models. The grading agents' confidence estimates and the interview agent's understanding labels require calibration against instructor judgments.

Interview coverage depends on question count and the prepared excerpts. Extraction follows text order, so prepared excerpts can overlap or omit later portions of long solutions. Written responses differ from the oral exchanges in human-led code interviews~\cite{kannam2025}, requiring attention to language accessibility.

Course deployment requires inference capacity and policies for access to and retention of student content across services. The workload analysis identifies request demand and ordering constraints; actual latency depends on capacity and contention at the HPC endpoint. Reproducible comparisons require model and skill versions, decoding settings, correction policies, and aggregation rules~\cite{hariri2026testtimescaling}. Browser activity reports alone cannot establish unauthorized assistance or justify a misconduct finding.

\section{Conclusion}
ProofPaths pairs assignment grading with an individualized reasoning interview to examine what students can explain about their solutions. Task-specific skills guide agents in rubric review and follow-up questioning, while instructors define assessment criteria and retain authority over grades. Source excerpts and prior answers give each interview its direction; retained reviews and transcripts make the resulting judgments inspectable. Institutional HPC resources supply inference, and versioned execution ties each result to its assessment inputs. ProofPaths automates recurring grading and interview tasks while preserving instructor control and an inspectable record of students' explanations.

\section*{Acknowledgment}
This work was supported in part by NSF awards 2117439 and 2320952 and used the HPC Core facility at CWRU.

We used GPT-6~\cite{openaigpt6} and a locally hosted Qwen3.8-27B model~\cite{qwen38model} for assistance with code, documentation, testing, figure preparation, and drafting and language polishing of the manuscript. The models also assisted with research ideation, literature search and synthesis, checks of implementation claims and references, and critical review. The authors take responsibility for the final manuscript and research artifacts, including all AI-assisted contributions.
\bibliographystyle{IEEEtran}
\bibliography{references,controls}

\end{document}